\documentclass[10pt,twocolumn,letterpaper]{article}

\usepackage{cvpr}
\usepackage{times}
\usepackage{epsfig}
\usepackage{graphicx}
\usepackage{amsmath}
\usepackage{amssymb}
\usepackage{color}
\usepackage{tabularx}
\usepackage{enumerate}
\usepackage{multirow}

\usepackage{bm}
\usepackage{verbatim}

\usepackage{adjustbox}

\newcommand{\Ads}{KWAI-AD}

\usepackage[breaklinks=true,bookmarks=false]{hyperref}

\cvprfinalcopy 

\def\cvprPaperID{4912} 

\begin{document}

\title{IMRAM: Iterative Matching with Recurrent Attention Memory \\ for Cross-Modal Image-Text Retrieval\thanks{This work was supported by the National Natural Science
Foundation of China (Nos. U1936202, 61925107). Corresponding author: Guiguang Ding}}

\author{Hui Chen\textsuperscript{1}, Guiguang Ding\textsuperscript{1*}, Xudong Liu\textsuperscript{2}, Zijia Lin\textsuperscript{3}, Ji Liu\textsuperscript{4}, Jungong Han\textsuperscript{5}\\
\textsuperscript{1}School of Software, BNRist, Tsinghua University\\
\textsuperscript{2}Kwai Ads Platform; \textsuperscript{3}Microsoft Research\\
\textsuperscript{4}Kwai Seattle AI Lab, Kwai FeDA Lab, Kwai AI Platform\\
\textsuperscript{5}WMG Data Science, University of Warwick\\
{\tt\small \{jichenhui2012, ji.liu.uwisc, jungonghan77\}@gmail.com}\\
{\tt\small dinggg@tsinghua.edu.cn, liuxudong@kuaishou.com, zijlin@microsoft.com}
}

\maketitle

\begin{abstract}
Enabling bi-directional retrieval of images and texts is important for understanding the correspondence between vision and language. Existing methods leverage the attention mechanism to explore such correspondence in a fine-grained manner. However, most of them consider all semantics equally and thus align them uniformly, regardless of their diverse complexities. In fact, semantics are diverse (\ie{} involving different kinds of semantic concepts), and humans usually follow a latent structure to combine them into understandable languages. It may be difficult to optimally capture such sophisticated correspondences in existing methods. 
In this paper, to address such a deficiency, we propose an Iterative Matching with Recurrent Attention Memory (IMRAM) method, in which correspondences between images and texts are captured with multiple steps of alignments. Specifically, we introduce an iterative matching scheme to explore such fine-grained correspondence progressively. A memory distillation unit is used to refine alignment knowledge from early steps to later ones. Experiment results on three benchmark datasets, \ie{} Flickr8K, Flickr30K, and MS COCO, show that our IMRAM achieves state-of-the-art performance, well demonstrating its effectiveness. Experiments on a practical business advertisement dataset, named \Ads{}, further validates the applicability of our method in practical scenarios.
\end{abstract}

\section{Introduction}
Due to the explosive increase of multimedia data from social media and web applications, enabling bi-directional cross-modal image-text retrieval is in great demand and
has become prevalent in both academia and industry. Meanwhile, this task is challenging because it requires to understand not only the content of images and texts but also their inter-modal correspondence~\cite{Karpathy2014Deep}.

In recent years, a large number of researches have been proposed and achieved great progress. Early works attempted to directly map the information of images and texts into a common latent embedding space. For example, Wang \etal~\cite{wang2016learning} adopted a deep network with two branches to, respectively, map images and texts into an embedding space. However, these works coarsely capture the correspondence between modalities and thus are unable to depict the fine-grained interactions between vision and language. 

To gain a deeper understanding of such fine-grained correspondences, recent researches further explored the attention mechanism for cross-modal image-text retrieval. Karpathy \etal{}~\cite{karpathy2015deep} extracted features of fragments for each image and text (\ie{} image regions and text words), and proposed a dense alignment between each fragment pair. Lee \etal{}~\cite{lee2018stacked} proposed a stacked cross attention model, in which attention was used to align each fragment with all fragments from another modality. It can neatly discover the fine-grained correspondence and thus achieves state-of-the-art performance on several benchmark datasets.

However, due to the large heterogeneity gap between images and texts, existing attention-based models, \eg{}~\cite{lee2018stacked}, may not well seize the optimal pairwise relationships among a number of region-word fragments pairs. Actually, semantics are complicated, because they are diverse (\ie{} composed by different kinds of semantic concepts with different meanings, such as objects (\eg nouns), attributes (\eg{} adjectives) and relations (\eg{} verbs)). And there generally exist strong correlations among different concepts, \eg{} relational terms (\eg{} verbs) usually indicate relationships between objects (\eg{} nouns). Moreover, humans usually follow a latent structure (\eg{} a tree-like structure~\cite{tai2015improved}) to combine different semantic concepts into understandable languages, which indicates that semantics shared between images and texts exhibit a complicated distribution. However, existing state-of-the-art models treat different kinds of semantics equally and align them together uniformly, taking little consideration of the complexity of semantics.

In reality, when humans perform comparisons between images and texts, we usually associate low-level semantic concepts, \eg{} objects, at the first glimpse. Then, higher-level semantics, \eg{} attributes and relationships, are mined by revisiting images and texts to obtain a better understanding~\cite{perlovsky2011language}. This intuition is favorably consistent with the aforementioned complicated semantics, and meanwhile, it indicates that the complicated correspondence between images and texts should be exploited progressively.

Motivated by this, in this paper, we propose an iterative matching framework with recurrent attention memory for cross-modal image-text retrieval, termed IMRAM. Our way of exploring the correspondence between images and texts is characterized by two main features: (1) an iterative matching scheme with a cross-modal attention unit to align fragments across different modalities; (2) a memory distillation unit to dynamically aggregate information from early matching steps to later ones. The iterative matching scheme can \textit{progressively} update the cross-modal attention core to accumulate cues for locating the matched semantics, while the memory distillation unit can refine the latent correspondence by enhancing the interaction of cross-modality information. Leveraging these two features, different kinds of semantics are treated distributively and well captured at different matching steps.

We conduct extensive experiments on several benchmark datasets for cross-modal image-text retrieval, \ie{} Flickr8K, Flickr30K, and MS COCO. Experiment results show that our proposed IMRAM can outperform the state-of-the-art models. Subtle analyses are also carried out to provide more insights about IMRAM. We observe that: (1) the fine-grained latent correspondence between images and texts can be well refined during the iterative matching process; (2) different kinds of semantics, respectively, play dominant roles at different matching steps in terms of contributions to the performance improvement. 

These observations can account for the effectiveness and reasonableness of our proposed method, which encourages us to validate its potential in practical scenarios. Hence, we collect a new dataset, named \Ads{}, by crawling about 81K image-text pairs on an advertisement platform, in which each image is associated with at least one advertisement textual title. We then evaluate our proposed method on the \Ads{} dataset and make comparisons with the state-of-the-art models. Results show that our method performs considerably better than compared models, further demonstrating the effectiveness of our method in the practical business advertisement scenario. The source code is available at: \url{https://github.com/HuiChen24/IMRAM}.

The contributions of our work are three folds: 1) First, we propose an iterative matching method for cross-modal image-text retrieval to handle the complexity of semantics. 2) Second, we formulate the proposed iterative matching method with a recurrent attention memory which incorporates a cross-modal attention unit and a memory distillation unit to refine the correspondence between images and texts. 3) Third, we verify our method on benchmark datasets (\ie{} Flickr8K, Flickr30K, and MS COCO) and a real-world business advertisement dataset (\ie{} our proposed \Ads{} dataset). Experimental results show that our method outperforms compared methods in all datasets. Thorough analyses on our model also well demonstrate the superiority and reasonableness of our method.

\begin{figure*}[!t] 
  \centering
  \includegraphics[width=0.95\linewidth]{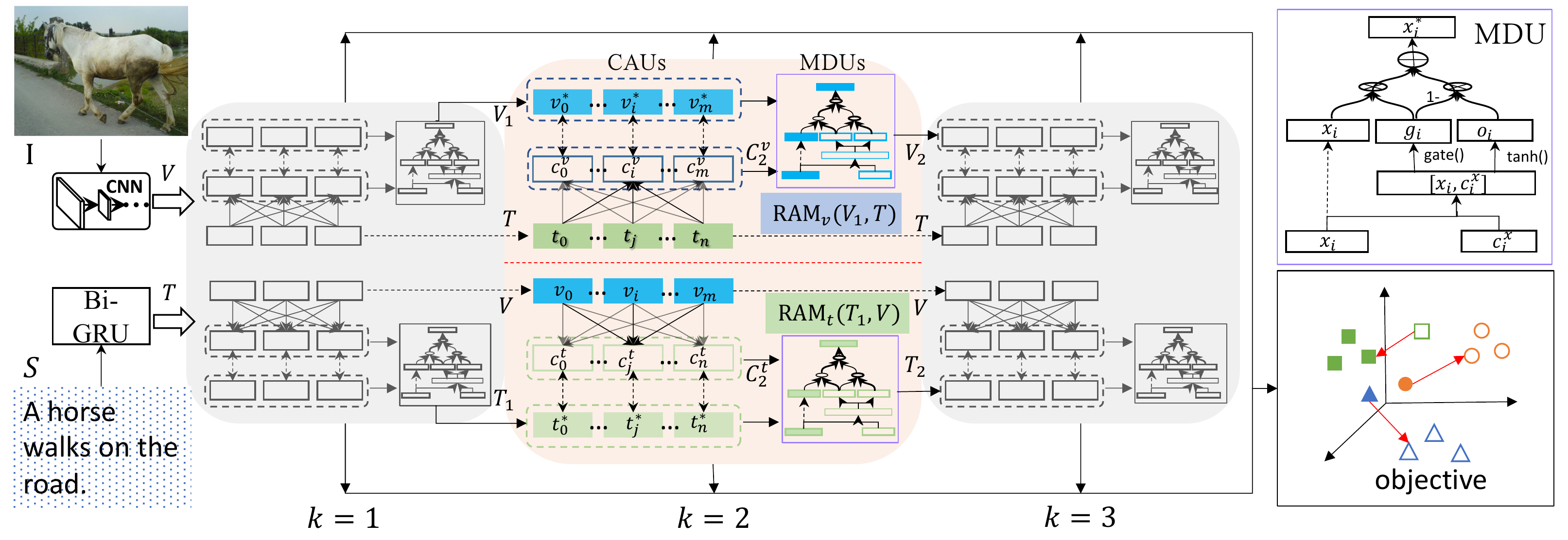}
  \caption{Framework of the proposed model.}
  \label{fig:framework}
\end{figure*}

\section{Related work}
Our work is concerned about the task of cross-modal image-text retrieval which essentially aims to explore the latent correspondence between vision and language. Existing matching methods can be roughly categorized into two lines: (1) coarse-grained matching methods aiming to mine the correspondence globally by mapping the whole images and the full texts into a common embedding space, (2) fine-grained matching ones aiming to explore the correspondence between image fragments and text fragments at a fine-grained level.


\textbf{Coarse-grained matching methods.} Wang \etal{}~\cite{wang2016learning} used a deep network with two branches of multilayer perceptrons to deal with images and texts, and optimized it with intra- and inter-structure preserving objectives. Kiros \etal{}~\cite{kiros2014unifying} adopted a CNN and a Gate Recurrent Unit (GRU) with a hinge-based triplet ranking loss to optimize the model by averaging the individual violations across the negatives. Alternatively, Faghri \etal{}~\cite{faghri2017vse++} reformed the ranking objective with a hard triplet loss function parameterized by only hard negatives.


\textbf{Fine-grained matching methods.} Recently, several works have been devoted to exploring the latent fine-grained vision-language correspondence for cross-modal image-text~\cite{karpathy2015deep,Niu2017Hierarchical,huang2017instance,nam2017dual,lee2018stacked}. 
Karpathy \etal ~\cite{karpathy2015deep} extracted features for fragments of each image and text, \ie{} image regions and text words, and aligned them in the embedding space. Niu \etal{}~\cite{Niu2017Hierarchical} organized texts as a semantic tree with each node corresponding to a phrase, and then used a hierarchical long short term memory (LSTM, a variant of RNN) to extract phrase-level features for text. 
Huang \etal{}~\cite{huang2017instance} presented a context-modulated attention scheme to selectively attend to salient pairwise image-sentence instances. Then a multi-modal LSTM was used to sequentially aggregate local similarities into a global one. Nam \etal{}~\cite{nam2017dual} proposed a dual attention mechanism in which salient semantics in images and texts were obtained by two attentions, and the similarity was computed by aggregating a sequence of local similarities. Lee \etal{}~\cite{lee2018stacked} proposed a stacked cross attention model which aligns each fragment with all other fragments from the other modality. They achieved state-of-the-art performance on several benchmark datasets for cross-modal retrieval.

While our method targets the same as~\cite{karpathy2015deep,lee2018stacked}, differently, we apply an \textit{iterative} matching scheme to refine the fragment alignment. Besides, we adopt a memory unit to distill the knowledge of matched semantics in images and texts after each matching step. Our method can also be regarded as a sequential matching method, as~\cite{nam2017dual,huang2017instance}. However, within the sequential computations, we transfer the knowledge about the fragment alignment to the successive steps with the proposed recurrent attention memory, instead of using modality-specific context information. Experiments also show that our method outperforms those mentioned works.

We also noticed that some latest works make use of large-scale external resources to improve performance. For example, Mithun \etal{}~\cite{Mithun2018Webly} collected amounts of image-text pairs from the Internet and optimized the retrieval model with them. Moreover, inspired by the recent great success of contextual representation learning for languages in the field of natural language processing (ELMO~\cite{Peters2018ELMO}, BERT~\cite{devlin2018bert} and XLNet~\cite{yang2019xlnet}), researchers also explored to apply BERT into cross-modal understanding field~\cite{Chen2019UNITERLU,li2019unicodervl}. However, such pre-trained cross-modal BERT models\footnote{Corresponding codes and models are not made publicly available.} require large amounts of annotated image-text pairs, which are not easy to obtain in the practical scenarios. On the contrary, our method is general and unlimited to the amount of data. We leave the exploration of large-scale external data to future works.

\section{Methodology}

In this section, we will elaborate on the details of our proposed IMRAM for cross-modal image-text retrieval. Figure~\ref{fig:framework} shows the framework of our model. We will first describe the way of learning the cross-modal feature representations in our work in section~\ref{sec:feature_representation}. Then, we will introduce the proposed recurrent attention memory as a module in our matching framework in section~\ref{sec:ram}. We will also present how to incorporate the proposed recurrent attention memory into the iterative matching scheme for cross-modal image-text retrieval in section~\ref{sec:iter_ram}. Finally, the objective function is discussed in section~\ref{sec:optimization}.

\subsection{Cross-modal Feature Representation}
\label{sec:feature_representation}
\textbf{Image representation.}
Benefiting from the development of deep learning in computer vision, different convolution neural networks have been widely used in many tasks to extract visual information for images. To obtain more descriptive information about the visual content for image fragments, we employ a pretrained deep CNN, \eg{} Faster R-CNN. Specifically, given an image $\bm{I}$, a CNN detects image regions and extracts a feature vector $\bm{f}_i$ for each image region $\bm{r}_i$. We further transform $\bm{f}_i$ to a $d$-dimensional vector $\bm{v_i}$ via a linear projection as follows:
\begin{equation}
\bm{v}_i = W_v \bm{f}_i + b_v
\label{equ:image_feature}
\end{equation}
where $W_v$ and $b_v$ are to-be-learned parameters.

For simplicity, we denote the image representation as $\bm{V} = \{\bm{v}_{i} | i=1,...,m, \bm{v}_{i} \in \mathbb{R}^d \}$, where $m$ is the number of detected regions in $\bm{I}$. We further normalize each region feature vector in $\bm{V}$ as~\cite{lee2018stacked}.

\textbf{Text representation.}
Basically, texts can be represented at either sentence-level or word-level. To enable the fine-grained connection of vision and language, we extract the word-level features for texts, which can be done through a bi-directional GRU as the encoder. 

Specifically, for a text $\bm{S}$ with $n$ words, we first represent each word $\bm{w}_j$ with a contiguous embedding vector $\bm{e}_j= \bm{W_e} \bm{w}_{j}, \forall j \in [1,n]$, where $\bm{W_e}$ is a to-be-learned embedding matrix. Then, to enhance the word-level representation with context information, we employ a bi-directional GRU to summarize information from both forward and backward directions in the text $\bm{S}$:
\begin{equation}
\label{equ:gru}
\begin{aligned}
& \overrightarrow{\bm{h}}_{j}=\overrightarrow{\bm{GRU}}(\bm{e}_{j}, \overrightarrow{\bm{h}}_{j-1}); \\
& \overleftarrow{\bm{h}}_{j}=\overleftarrow{\bm{GRU}}(\bm{e}_{j}, \overleftarrow{\bm{h}}_{j+1})
\end{aligned}
\end{equation}
where $\overrightarrow{\bm{h}}_{j}$ and $\overleftarrow{\bm{h}}_{j}$ denote hidden states from the forward GRU and the backward GRU, respectively. Then, the representation of the word $\bm{w}_j$ is defined as $\bm{t}_{j} = \frac{\overrightarrow{\bm{h}}_{j}+\overleftarrow{\bm{h}}_{j}}{2}.$

Eventually, we obtain a word-level feature set for the text $\bm{S}$, denoted as $\bm{T} = \{\bm{t}_{j} | j=1,...,n, \bm{t}_{j} \in \mathbb{R}^d \}$, where each $\bm{t}_{j}$ encodes the information of the word $\bm{w}_j$. Note that each $\bm{t}_j$ shares the same dimensionality as $\bm{v}_{i}$ in Eq. \ref{equ:image_feature}. We also normalize each word feature vector in $\bm{T}$ as \cite{lee2018stacked}.

\subsection{RAM: Recurrent Attention Memory}
\label{sec:ram}
The recurrent attention memory aims to align fragments in the embedding space by refining the knowledge about previous fragment alignments in a recurrent manner. It can be regarded as a block that takes in two sets of feature points, \ie{} $\bm{V}$ and $\bm{T}$, and estimates the similarity between these two sets via a cross-modal attention unit. A memory distillation unit is used to refine the attention result in order to provide more knowledge for the next alignments. For generalization, we denote the two input sets of features as a query set $\bm{X}=\{\bm{x}_i|i \in [1,m'], \bm{x}_{i} \in \mathbb{R}^d\}$ and a response set $\bm{Y}=\{\bm{y}_j|j\in[1,n'], \bm{y}_{j} \in \mathbb{R}^d\}$, where $m'$ and $n'$ are the numbers of feature points in $\bm{X}$ and $\bm{Y}$, respectively. Note that $\bm{X}$ can be either of $\bm{V}$ and $\bm{T}$, while $\bm{Y}$ will be the other. 

\textbf{Cross-modal Attention Unit (CAU).}
The cross-modal attention unit aims to summarize context information in $\bm{Y}$ for each feature $\bm{x}_i$ in $\bm{X}$. To achieve this goal, we first compute the similarity between each pair $(\bm{x}_i, \bm{y}_j)$ using the cosine function:
\begin{equation}
z_{ij}=\frac{\bm{x}_i^T \bm{y}_j}{|| \bm{x}_i || \cdot || \bm{y}_j ||}, \forall i \in [1,m'], \forall j \in [1,n']
\label{equ:cosine_similarity}
\end{equation}
As \cite{lee2018stacked}, we further normalize the similarity score $z$ as: 
\begin{equation}
\bar{z}_{ij}=\frac{\text{relu}(z_{ij})}{\sqrt{\sum_{i=1}^{m'} \text{relu}(z_{ij})^2}}
\label{equ:region_norm}
\end{equation}
where $\text{relu}(x)=\text{max}(0,x)$.

Attention is performed over the response set $\bm{Y}$ given a feature $\bm{x}_i$ in $\bm{X}$:
\begin{equation}
\begin{aligned}
& \bm{c}_i^x = \sum_{j=1}^{n'} \alpha_{ij} \bm{y}_j,
& s.t. \quad \alpha_{ij} = \frac{exp(\lambda \bar{z}_{ij})}{\sum_{j=1}^{n'} exp(\lambda \bar{z}_{ij})}
\label{equ:attention}
\end{aligned}
\end{equation}
where $\lambda$ is the inverse temperature parameter of the softmax function~\cite{chorowski2015attention} to adjust the smoothness of the attention distribution.

We define $\bm{C}^x=\{\bm{c}_i^x | i \in [1,m'], \bm{c}^{x}_i \in \mathbb{R}^d\}$ as $\bm{X}$-grounded alignment features, in which each element captures related semantics shared by each $\bm{x}_i$ and the whole $\bm{Y}$.

\textbf{Memory Distillation Unit (MDU).}
To refine the alignment knowledge for the next alignment, we adopt a memory distillation unit which updates the query features $\bm{X}$ by aggregating them with the corresponding $\bm{X}$-grounded alignment feature $\bm{C}^x$ dynamically:
\begin{equation}
\bm{x}^{*}_i = f(\bm{x}_i, \bm{c}_i^x)
\label{equ:mem_distill}
\end{equation}
where $f()$ is a aggregating function. We can define $f()$ with different formulations, such as addition, multilayer perceptron (MLP), attention and so on. Here, we adopt a modified gating mechanism for $f()$:
\begin{equation}
\label{equ:memory}
\begin{aligned}
& \bm{g}_i=\text{gate}(W_g[\bm{x}_i, \bm{c}_i^x] + b_g) \\
& \bm{o}_i=\text{tanh}(W_o[\bm{x}_i, \bm{c}_i^x] + b_o) \\
& \bm{x}^{*}_i=\bm{g}_i * \bm{x}_i + (1-\bm{g}_i) * \bm{o}_i \\
\end{aligned}
\end{equation}
where $W_g, W_o, b_g, b_o$ are to-be-learned parameters. $\bm{o}_i$ is a fused feature which enhances the interaction between $\bm{x}_i$ and $\bm{c}_i^x$. $\bm{g}_i$ performs as a gate to select the most salient information.

With the gating mechanism, information of the input query can be refined by itself (\ie{} $\bm{x}_i$) and the semantic information shared with the response (\ie{} $\bm{o}_i$). The gate $\bm{g}_i$ can help to filter trivial information in the query, and enable the representation learning of each query fragment (\ie{} $\bm{x}_i$ in $\bm{X}$) to focus more on its individual shared semantics with $\bm{Y}$. Besides, the $\bm{X}$-grounded alignment features $\bm{C}^x$ summarize the context information of $\bm{Y}$ with regard to each fragment in $\bm{X}$. And in the next matching step, such context information will assist to determine the shared semantics with respect to $\bm{Y}$, forming a recurrent computation process as described in the subsequent section~\ref{sec:iter_ram}. Therefore, with the help of $\mathbf{C}^x$, the intra-modality relationships in $\bm{Y}$ are implicitly involved and re-calibrated during the recurrent process, which would enhance the interaction among cross-modal features and thus benefit the representation learning.

\textbf{RAM block.} We integrate the cross-modal attention unit and the memory distillation unit into a RAM block, formulated as:
\begin{equation}
\bm{C}^x, \bm{X}^{*} = \textbf{RAM}(\bm{X}, \bm{Y})
\label{equ:ram}
\end{equation}
where $\bm{C}^x$ and $\bm{X}^{*}$ are derived by Eq. \ref{equ:attention} and \ref{equ:mem_distill}.

\begin{table*}[t!]
\centering
\caption{Comparison with the state-of-the-art models on Flickr8K. As results of SCAN~\cite{lee2018stacked} are not reported on Flickr8K, here we show our experiment results by running codes provided by authors.}
\label{tab:flickr8k}
\setlength{\tabcolsep}{6mm}{
\begin{adjustbox}{max width=0.9\textwidth}
    \scalebox{1.0}{
  \begin{tabular}{l|cccccc|c}
    \hline
    \multirow{2}{*}{Method} &
    \multicolumn{3}{c}{Text Retrieval} &
    \multicolumn{3}{c|}{Image Retrieval} &
    \multirow{2}{*}{R@sum}\\
    & R@1 & R@5 & R@10 & R@1 & R@5 & R@10  & \\
    \hline
    DeViSE~\cite{frome2013devise} & 4.8 & 16.5 & 27.3 & 5.9 & 20.1 & 29.6 &104.2\\
    DVSA~\cite{karpathy2015deep} & 16.5 & 40.6 & 54.2 &11.8 &32.1 &44.7 &199.9\\
    m-CNN~\cite{ma2015multimodal} & 24.8 & 53.7 & 67.1 & 20.3 &47.6 & 61.7 &275.2\\
    SCAN* &52.2 &81.0  &89.2 &38.3 &67.8 &78.9 &407.4 \\
    \hline
    Image-IMRAM &48.5 &78.1 &85.3 &32.0 &61.4 &73.9 &379.2 \\
    Text-IMRAM &52.1 &81.5 &90.1 &40.2 &69.0 &79.2 &412.1\\
    Full-IMRAM &\textbf{54.7}&\textbf{84.2}&\textbf{91.0}&\textbf{41.0}&\textbf{69.2}&\textbf{79.9} &\textbf{420.0}\\
    \hline
  \end{tabular}}
\end{adjustbox}}
\end{table*}

\subsection{Iterative Matching with Recurrent Attention Memory}
\label{sec:iter_ram}
In this section, we describe how to employ the recurrent attention memory introduced above to enable the iterative matching for cross-modal image-text retrieval.

Specifically, given an image $\bm{I}$ and a text $\bm{S}$, we derive two strategies for iterative matching grounded on $\bm{I}$ and $\bm{S}$, respectively, using two independent RAM blocks:
\begin{equation}
\begin{aligned}
& \bm{C}^v_{k}, \bm{V}_{k}=\textbf{RAM}_v(\bm{V}_{k-1}, \bm{T}) \\
& \bm{C}^t_{k}, \bm{T}_{k}=\textbf{RAM}_t(\bm{T}_{k-1}, \bm{V}) \\
\end{aligned}
\end{equation}
where $\bm{V}_k$, $\bm{T}_k$ indicate the step-wise features of the image $\bm{I}$ and the text $\bm{S}$, respectively. And $k$ is the matching step, and $\bm{V}_{0}=\bm{V}$, $\bm{T}_{0}=\bm{T}$.

We iteratively perform \textbf{RAM()} for a total of $K$ steps. And at each step $k$, we can derive a matching score between $\bm{I}$ and $\bm{S}$:
\begin{equation}
\label{equ:iterative_match}
F_k(\bm{I},\bm{S})=\frac{1}{m}\sum_{i=1}^m F_k(\bm{r}_i,\bm{S})+ \frac{1}{n}\sum_{j=1}^n F_k(\bm{I}, \bm{w}_j)
\end{equation}
where $F(\bm{r}_i,\bm{S})$ and $F(\bm{I}, \bm{w}_j)$ are defined as the region-based matching score and the word-based matching score, respectively. They are derived as follows:

\begin{equation}
\label{equ:word-region-similarity}
\begin{aligned}
& F_k(\bm{r}_i, \bm{S})=\text{sim}(\bm{v}_i, \bm{c}^v_{ki}); \\
& F_k(\bm{I}, \bm{w}_j)=\text{sim}(\bm{c}^t_{kj}, \bm{t}_j) 
\end{aligned}
\end{equation}
where \text{sim()} is the cosine function that measures the similarity between two input features as Eq. \ref{equ:cosine_similarity}. And $\bm{v}_i \in \bm{V}$ corresponds to the region $\bm{r}_i$. $\bm{t}_j \in \bm{T}$ corresponds to the word $\bm{w}_j$. $\bm{c}^v_{ki} \in \bm{C}^v_k$ and $ \bm{c}^t_{kj} \in \bm{C}^t_k$ are, respectively, the context feature corresponding to the region $\bm{r}_i$ and the word $\bm{w}_j$. $m$ and $n$ are the numbers of image regions and text words, respectively.

After $K$ matching steps, we derive the similarity between $\bm{I}$ and $\bm{S}$ by summing all matching scores:
\begin{equation}
F(\bm{I},\bm{S}) = \sum_{k=1}^K F_k(\bm{I},\bm{S})
\label{equ:final_score}
\end{equation}

\subsection{Loss Function}
\label{sec:optimization}
In order to enforce matched image-text pairs to be clustered and unmatched ones to be separated in the embedding spaces, triplet-wise ranking objectives are widely used in previous works~\cite{kiros2014unifying,faghri2017vse++} to train the model in an end-to-end manner. 
Following \cite{faghri2017vse++}, instead of comparing with all negatives, we only consider the \textit{hard} negatives within a mini-batch, \ie{} the negative that is closest to a training query:
\begin{equation}
\begin{aligned}
\mathcal{L} = \sum_{b=1}^{B}[\Delta - F(I_b, S_b) + F(I_b, S_{b^*})]_{+} \\
 + \sum_{b=1}^{B}[\Delta - F(I_b, S_b) + F(I_{b^*}, S_b)]_{+} \\
 \end{aligned}
\label{equ:ranking}
\end{equation}
where $[x]_{+}=max(x,0)$, and $F(I,S)$ is the semantic similarity between $I$ and $S$ defined by Eq. \ref{equ:final_score}. Images and texts with the same subscript $b$ are matched examples. Hard negatives are indicated by the subscript $b^*$. $\Delta$ is a margin value.

Note that in the loss function, $F(\bm{I}, \bm{S})$ consists of $F_k(\bm{I}, \bm{S})$ at each matching step (\ie{} Eq. \ref{equ:final_score}), and thus optimizing the loss function would directly supervise the learning of image-text correspondences at each matching step, which is expected to help the model to yield higher-quality alignment at each step. With the employed triplet-wise ranking objective, the whole model parameters can be optimized in an end-to-end manner, using widely-used optimizers like SGD, etc.


\begin{table*}[t!]
\centering
\caption{Comparison with state-of-the-art models on Flickr30K.}
\label{tab:flickr30k}
\setlength{\tabcolsep}{6mm}{
\begin{adjustbox}{max width=0.9\textwidth}
    \scalebox{1}{
  \begin{tabular}{l|cccccc|c}
    \hline
    \multirow{2}{*}{Method} &
    \multicolumn{3}{c}{Text Retrieval} &
    \multicolumn{3}{c|}{Image Retrieval} &
    \multirow{2}{*}{R@sum} \\
    & R@1 & R@5 & R@10 & R@1 & R@5 & R@10 & \\
    \hline
    DPC~\cite{zheng2017dual} & 55.6 & 81.9 & 89.5 & 39.1 & 69.2 & 80.9 &416.2\\
    SCO~\cite{huang2018learning} & 55.5 & 82.0 & 89.3 & 41.1 & 70.5 & 80.1 &418.5\\
    SCAN*~\cite{lee2018stacked} &67.4 & 90.3 & 95.8 & 48.6 & 77.7 & 85.2 &465.0\\
    VSRN*~\cite{li2019visual} &71.3 &90.6 &96.0 &\textbf{54.7} &\textbf{81.8} &\textbf{88.2} & 482.6 \\
    \hline
    Image-IMRAM &67.0 &90.5 &95.6 &51.2 &78.2 &85.5 &468.0\\
    Text-IMRAM &68.8 &91.6 &96.0 &53.0 &79.0 &87.1 &475.5 \\
    Full-IMRAM &\textbf{74.1}&\textbf{93.0}&\textbf{96.6}&\textbf{53.9}&\textbf{79.4}&\textbf{87.2} &\textbf{484.2}\\
    \hline
  \end{tabular}}
\end{adjustbox}}
\end{table*}

\begin{table*}[t!]
\centering
\caption{Comparison with state-of-the-art models on MS COCO.}
\label{tab:ms_coco}
\setlength{\tabcolsep}{6mm}{
\begin{adjustbox}{max width=0.9\textwidth}
    \scalebox{1}{
  \begin{tabular}{l|cccccc|c}
    \hline
    \multirow{2}{*}{Method} &
    \multicolumn{3}{c}{Text Retrieval} &
    \multicolumn{3}{c|}{Image Retrieval} &
    \multirow{2}{*}{R@sum}\\
    & R@1 & R@5 & R@10 & R@1 & R@5 & R@10 & \\
    \hline
    \multicolumn{8}{c}{1K} \\
    \hline
    DPC~\cite{zheng2017dual} & 65.6 & 89.8 & 95.5 & 47.1 & 79.9 & 90.0 &467.9\\
    SCO~\cite{huang2018learning} & 69.9 & 92.9 & 97.5 & 56.7 & 87.5 &94.8 &499.3\\
    SCAN*~\cite{lee2018stacked} &72.7 & 94.8 &98.4 & 58.8 & 88.4 & 94.8 &507.9\\
    PVSE~\cite{song2019Polysemous} &69.2 &91.6 &96.6 &55.2 &86.5 &93.7  &492.8 \\
    VSRN*~\cite{li2019visual} &76.2  &94.8  &98.2  &\textbf{62.8}  &\textbf{89.7}  &\textbf{95.1} &516.8 \\
    \hline
    Image-IMRAM &76.1  &95.3 &98.2 &61.0  &88.6 &94.5 &513.7 \\
    Text-IMRAM &74.0 &95.6 &98.4 &60.6 &88.9 &94.6 &512.1 \\
    Full-IMRAM &\textbf{76.7}&\textbf{95.6}&\textbf{98.5}&\textbf{61.7}&\textbf{89.1}&\textbf{95.0} &\textbf{516.6}\\
    \hline
    \multicolumn{8}{c}{5K} \\
    \hline
    DPC~\cite{zheng2017dual} & 41.2 & 70.5 & 81.1 &25.3 & 53.4 &66.4 &337.9\\
    SCO~\cite{huang2018learning} & 42.8 & 72.3 & 83.0 & 33.1 & 62.9 & 75.5 &369.6\\
    SCAN*~\cite{lee2018stacked} & 50.4 &82.2 &90.0 &38.6 &69.3 &80.4 &410.9 \\
    PVSE~\cite{song2019Polysemous} &45.2 &74.3 &84.5 &32.4 &63.0 &75.0 &374.4\\
    VSRN*~\cite{li2019visual} &53.0 &81.1 &89.4 &\textbf{40.5} &\textbf{70.6} &\textbf{81.1} &415.7\\
    \hline
    Image-IMRAM &53.2  &82.5 &90.4 &38.9 &68.5 &79.2 &412.7 \\
    Text-IMRAM &52.0 &81.8 &90.1 &38.6 &68.1 &79.1 &409.7 \\
    Full-IMRAM &\textbf{53.7}&\textbf{83.2}&\textbf{91.0}&\textbf{39.7}&\textbf{69.1}&\textbf{79.8} &\textbf{416.5}\\
    \hline
  \end{tabular}}
\end{adjustbox}}
\end{table*}

\section{Experiment}

\subsection{Datasets and Evaluation Metric}
\label{sec:exp_dataset}

Three benchmark datasets are used in our experiments, including: (1) \textbf{Flickr8K}: contains 8,000 images and provides 5 texts for each image. We adopt its standard splits as \cite{Niu2017Hierarchical,ma2015multimodal}, using 6,000 images for training, 1,000 images for validation and another 1,000 images for testing. (2) \textbf{Flickr30K}: consists of 31,000 images and 158,915 English texts. Each image is annotated with 5 texts. We follow the dataset splits as \cite{lee2018stacked,faghri2017vse++} and use 29,000 images for training, 1,000 images for validation, and the remaining 1,000 images for testing. (3) \textbf{MS COCO}: is a large-scale image description dataset containing about 123,287 images with at least 5 texts for each. As previous works~\cite{lee2018stacked,faghri2017vse++}, we use 113,287 images to train all models, 5,000 images for validation and another 5,000 images for testing. Results on MS COCO are reported by averaging over 5 folds of 1K test images and testing on the full 5K test images as~\cite{lee2018stacked}.

\begin{figure}[!t]
  \centering
 \includegraphics[width=0.9\linewidth]{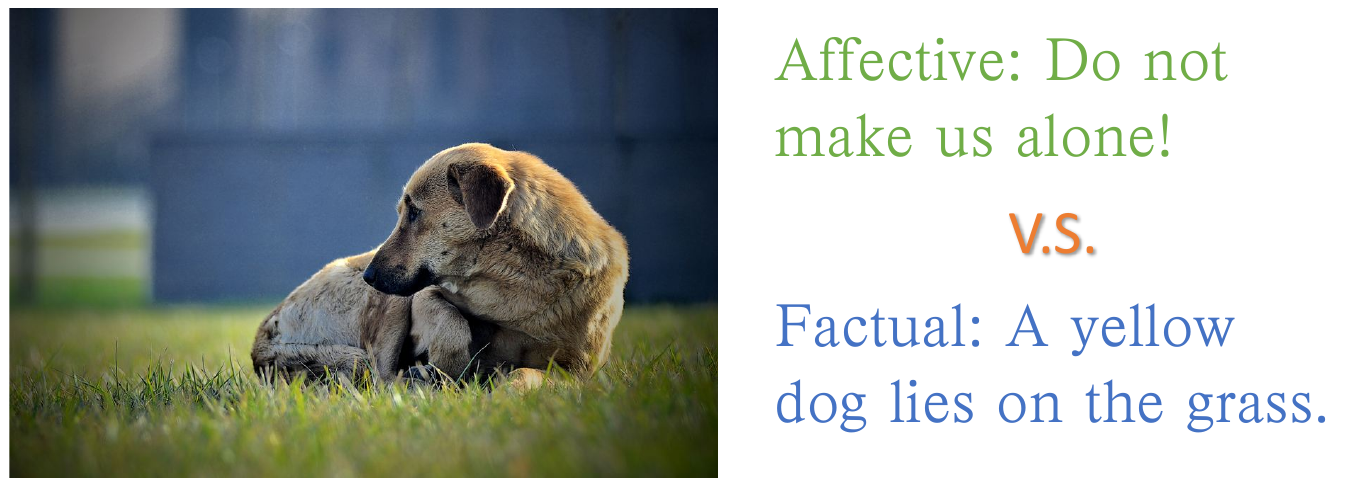}
  \caption{Difference between our \Ads{} dataset and standard datasets, \eg{} MS COCO.}
  \label{fig:weakly_ads}
\end{figure}

To further validate the effectiveness of our method in practical scenarios, we build a new dataset, named \textbf{\Ads{}}. We collect 81,653 image-text pairs from a real-world business advertisement platform, and we randomly sample 79,653 image-text pairs for training, 1,000 for validation and the remaining 1,000 for testing. The uniqueness of our dataset is that the provided texts are not detailed textual descriptions of the content in the corresponding images, but maintain weakly associations with them, conveying strong affective semantics instead of factual semantics (seeing Figure~\ref{fig:weakly_ads}). And thus our dataset is more challenging than conventional datasets. However, it is of great importance in the practical business scenario. Learning subtle links of advertisement images with related well-designed titles could not only enrich the understanding of vision and language but also benefit the development of recommender systems and social networks.

\textbf{Evaluation Metric.} To compare our proposed method with the state-of-the-art methods, we adopt the same evaluation metrics in all datasets as~\cite{Mithun2018Webly,lee2018stacked,faghri2017vse++}. Namely, we adopt Recall at K (R@K) to measure the performance of bi-directional retrieval tasks, \ie{} retrieving texts given an image query (Text Retrieval) and retrieving images given a text query (Image Retrieval). We report R@1, R@5, and R@10 for all datasets as in~\cite{lee2018stacked}. And to well reveal the effectiveness of the proposed method, we also report an extra metric ``R@sum'', which is the summation of all evaluation metrics as~\cite{huang2017instance}.

\subsection{Implementation Details}
\label{sec:exp_imple}
To systematically validate the effectiveness of the proposed IMRAM, we experiment with three of its variants: (1) Image-IMRAM only adopts the RAM block grounded on images (\ie{} only using the first term in Eq.~\ref{equ:iterative_match}); (2) Text-IMRAM only adopts the RAM block grounded on texts (\ie{} only using the first term in Eq.~\ref{equ:iterative_match}); (3) Full-IMRAM.
All models are implemented by Pytorch v1.0. In all datasets, for each word in texts, the word embedding is initialized by random weights with a dimensionality of 300. We use a bi-directional GRU with one layer and set its hidden state (\ie{} $\overrightarrow{\bm{h}}_{j}$ and $\overleftarrow{\bm{h}}_{j}$ in Eq. \ref{equ:gru}) dimensionality as 1,024. The dimensionality of each region feature (\ie{} $\bm{v}_i$ in $\bm{V}$) and and each word feature (\ie{} $\bm{t}_j$ in $\bm{T}$) is set as 1,024. On three benchmark datasets, we use Faster R-CNN pre-trained on Visual Genome to extract 36 region features for each image. For our \Ads{} dataset, we simply use Inception v3~\cite{szegedy2016rethinking} to extract 64 features for each image.

\subsection{Results on Three Benchmark Datasets}
\label{sec:exp_compare}
We compare our proposed IMRAM with published state-of-the-art models in the three benchmark datasets\footnote{We omit models that require additional data augmentation~\cite{Nguyen2019Multi-Task,Shi2019Knowledge,Mithun2018Webly,li2019unicodervl,Chen2019UNITERLU,ji2019saliency}.}. We directly cite the best-reported results from respective papers when available. And for our proposed models, we perform $3$ steps of iterative matching by default.

\textbf{Results.} Comparison results are shown in Table~\ref{tab:flickr8k}, Table~\ref{tab:flickr30k} and Table~\ref{tab:ms_coco} for Flickr8K, Flickr30K and MS COCO, respectively. `*' indicates the performance of an ensemble model. `-' means unreported results. We can see that our proposed IMRAM can consistently achieve performance improvements in terms of all metrics, compared to the state-of-the-art models. 

Specifically, our Full-IMRAM can significantly outperform the previous best model, \ie{} SCAN*~\cite{lee2018stacked}, by a large margin of $12.6\%$, $19.2\%$, $8.7\%$ and $5.6\%$ in terms of the overall performance R@sum in Flickr8K, Flickr30K, MS COCO (1K) and MS COCO (5K), respectively. And among recall metrics for the text retrieval task, our Full-IMRAM can obtain a maximal performance improvement of $3.2\%$ (R@5 in Flickr8K), $6.7\%$ (R@1 in Flickr30K), $4.0\%$ (R@1 in MS COCO (1K)) and $3.3\%$ (R@1 in MS COCO (5K)), respectively. As for the image retrieval task, the maximal improvements are $2.7\%$ (R@1 in Flickr8K), $5.3\%$ (R@1 in Flickr30K), $2.9\%$ (R@1 in MS COCO (1K)) and $1.1\%$ (R@1 in MS COCO (5K)), respectively. These results well demonstrate that the proposed method exhibits great effectiveness for cross-modal image-text retrieval. Besides, our models can consistently achieve state-of-the-art performance not only in small datasets, \ie{} Flickr8K and Flickr30K, but also in the large-scale dataset, \ie{} MS COCO, which well demonstrates its robustness.

\begin{table}
\centering
\caption{The effect of the total steps of matching, $K$, on variants of IMRAM in MS COCO (5K). }
\label{tab:matching_steps_coco}
\setlength{\tabcolsep}{1.5mm}{
\begin{tabular}{l|l|cccc}
  \hline
  \multirow{2}{*}{Model} &
  \multirow{2}{*}{$K$} &
  \multicolumn{2}{c}{Text Retrieval} &
  \multicolumn{2}{c}{Image Retrieval} \\
  & & R@1 & R@10 & R@1 & R@10\\
  \hline
  \multirow{3}{*}{Image}
  & 1 &40.8 &85.7 &34.6 &76.2 \\
  & 2 &51.5 &89.5 &37.7 &78.3 \\
  -IMRAM& 3 &\textbf{53.2} &\textbf{90.4} &\textbf{38.9} &\textbf{79.2} \\
  \hline
  \multirow{3}{*}{Text}
  & 1 &46.2 &87.0 &34.4 &75.9 \\
  & 2 &50.4 &89.2 &37.4 &78.3\\
  -IMRAM& 3 &\textbf{51.4} &\textbf{89.9} &\textbf{39.2} &\textbf{79.2} \\
  \hline
  \multirow{3}{*}{Full}
  & 1 &49.7 &88.9 &35.4 &76.7 \\
  & 2 &53.1 &90.2 &39.1 &79.5 \\
  -IMRAM& 3 &\textbf{53.7} &\textbf{91.0} &\textbf{39.7} &\textbf{79.8} \\
  \hline
\end{tabular}}
\end{table}

\subsection{Model Analysis}
\textbf{Effect of the total steps of matching, $K$.}
For all three variants of IMRAM, we gradually increase $K$ from 1 to 3 to train and evaluate them on the benchmark datasets. Due to the limited space, we only report results on MS COCO (5K test) in Table~\ref{tab:matching_steps_coco}. We can observe that for all variants, $K=2$ and $K=3$ can consistently achieve better performance than $K=1$. And $K=3$ performs better or comparatively, compared with $K=2$. This observation well demonstrates that the iterative matching scheme effectively improves model performance. Besides, our Full-IMRAM consistently outperforms Image-IMRAM and Text-IMRAM for different values of $K$.


\textbf{Effect of the memory distillation unit.} The aggregation function $f(\bm{x},\bm{y})$ in Eq. \ref{equ:mem_distill} is essential for the proposed iterative matching process. We enumerate some basic aggregation functions and compare them with ours: (1) \textsf{add}: $\bm{x}+\bm{y}$; (2) \textsf{mlp}: $\bm{x}+\text{tanh}(W\bm{y}+b)$; (3) \textsf{att}: $\alpha\bm{x} + (1-\alpha)\bm{y}$ where $\alpha$ is a real-valued number parameterized by $\bm{x}$ and $\bm{y}$; (4) \textsf{gate}: $\bm{\beta} \bm{x} + (1-\bm{\beta})\bm{y}$ where $\bm{\beta}$ is a real-valued vector parameterized by $\bm{x}$ and $\bm{y}$. We conduct the analysis with Text-IMRAM ($K=3$) in Flickr30K in Table~\ref{tab:memory_flickr}. We can observe that the aggregation function we use (\ie{} Eq. \ref{equ:memory}) achieves substantially better performance than baseline functions.



\begin{table}
\centering
\caption{The effect of the aggregating function in the proposed memory distillation unit of Text-IMRAM ($K=3$) in Flickr30K.}
\label{tab:memory_flickr}
\begin{tabular}{l|cccc}
  \hline
  \multirow{2}{*}{Memory} &
  \multicolumn{2}{c}{Text Retrieval} &
  \multicolumn{2}{c}{Image Retrieval} \\
  & R@1 & R@10 & R@1 & R@10\\
  \hline
  \textsf{add} &64.5 &95.1 &49.2 &84.9 \\
  \textsf{mlp} &66.6 &\textbf{96.4} &52.8 &86.2 \\
  \textsf{att} &66.1 &95.5 &52.1 &86.2 \\
  \textsf{gate} &66.2 &\textbf{96.4} &52.5 &86.1 \\
  \textsf{ours} &\textbf{68.8} &96.0 &\textbf{53.0} &\textbf{87.1}\\
  \hline
\end{tabular}
\end{table}

\begin{table}
\centering
\caption{Statistical results of salient semantics at each matching step, $k$, in Text-IMRAM ($K=3$) in MS COCO.}
\label{tab:statistical_hierarchy_k}
\begin{tabular}{c|c|c|c}
  \hline
  $k$ & nouns (\%) & verbs (\%) & adjectives (\%) \\
  \hline
  1 & 99.0 & 32.0 & 35.3 \\
  2 & 99.0 & 38.8 & 37.9 \\
  3 & \textbf{99.0} & \textbf{40.2} & \textbf{39.1} \\
  \hline
\end{tabular}
\end{table}

\begin{figure*}[!t]
  \centering
 \includegraphics[width=0.95\linewidth]{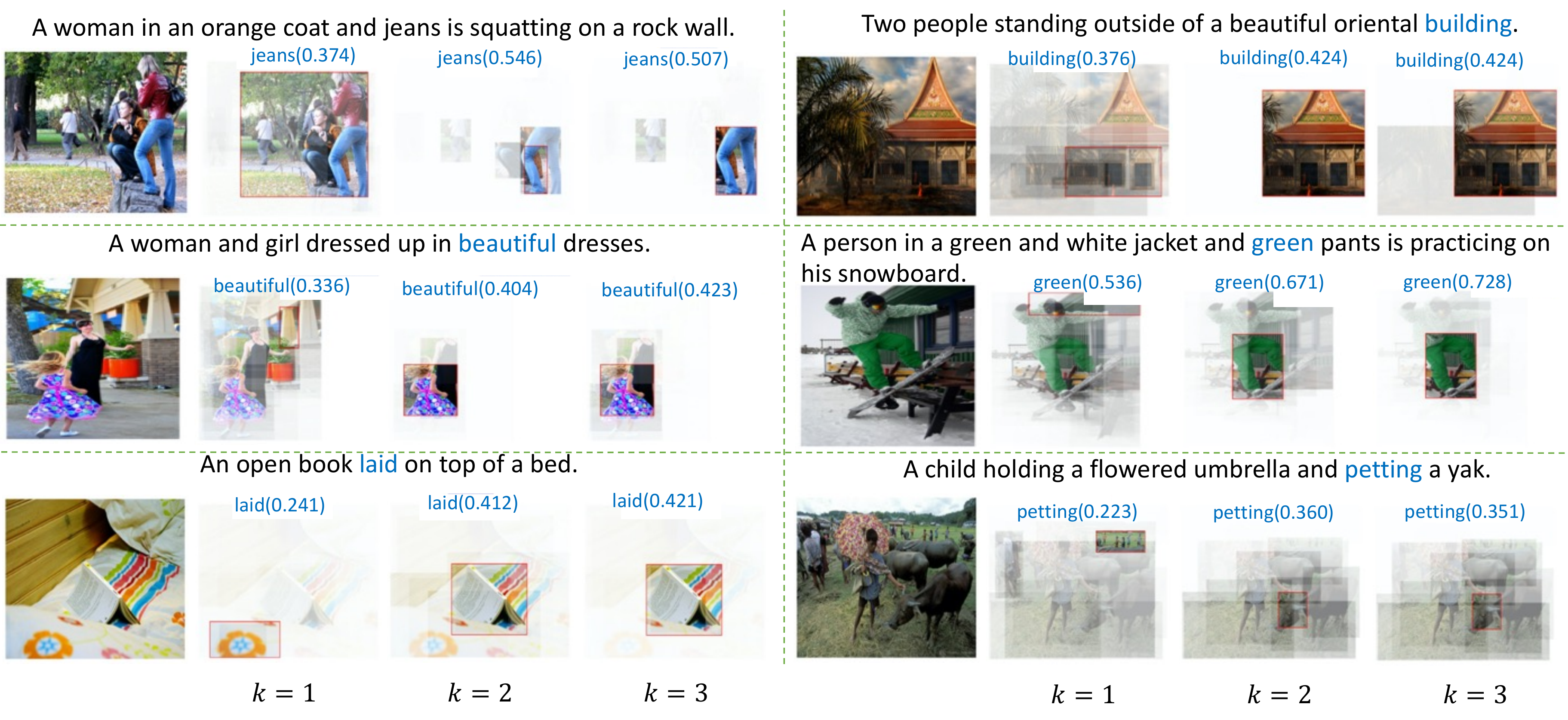}
  \caption{Visualization of attention at each matching step in Text-IMRAM. Corresponding matched words are in blue, followed by the matching similarity.}
  \label{fig:heirarchy_attention}
\end{figure*}

\subsection{Qualitative Analysis}
We intend to explore more insights for the effectiveness of our models here. For the convenience of the explanation, we mainly analyze semantic concepts from the view of language, instead of from the view of vision, \ie{} we treat each word in the text as one semantic concept. Therefore, we conduct the qualitative analysis on Text-IMRAM.

We first visualize the attention map at each matching step in Text-IMRAM ($K=3$) corresponding to different semantic concepts in Figure \ref{fig:heirarchy_attention}. We can see that the attention is refined and gradually focuses on the matched regions.

To quantitatively analyze the alignment of semantic concepts, we first define a semantic concept in Text-IMRAM as a salient one at the matching step $k$ as follows: 1) Given an image-text pair, at the matching step $k$, we derive the word-based matching score by Eq. \ref{equ:word-region-similarity} for each word with respect to the image, and derive the image-text matching score by averaging all the word-based scores (see Eq. \ref{equ:iterative_match}). 2) A semantic concept is salient if its corresponding word-based score is greater than the image-text score. For a set of image-text pairs randomly sampled from the testing set, we can compute the percentage of such salient semantic concepts for each model at different matching steps.

Then we analyze the change of the salient semantic concepts captured at different matching steps in Text-IMRAM ($K=3$). Statistical results are shown in Table~\ref{tab:statistical_hierarchy_k}. We can see that at the 1st matching step, nouns are easy to be recognized and dominant to help to match. While during the subsequent matching steps, contributions of verbs and adjectives increase.

\begin{table}
\centering
\caption{Results on the Ads dataset.}
\label{tab:ads_res}
\setlength{\tabcolsep}{1.5mm}{
\begin{tabular}{l|cccc}
  \hline
  \multirow{2}{*}{Method} &
  \multicolumn{2}{c}{Text Retrieval} &
  \multicolumn{2}{c}{Image Retrieval} \\
  &  R@1 & R@10 & R@1 & R@10\\
  \hline
  i-t AVG~\cite{lee2018stacked}&7.4 &21.1 &2.1 &9.3 \\
  Image-IMRAM&\textbf{10.7} &\textbf{25.1} &\textbf{3.4} &\textbf{16.8}  \\
   
  \hline
  t-i AVG~\cite{lee2018stacked} &6.8 &20.8 & 2.0 & 9.9 \\
  Text-IMRAM&\textbf{8.4} &\textbf{21.5} &\textbf{2.3} &\textbf{15.9} \\
  \hline
  i-t + t-i~\cite{lee2018stacked} &7.3 &22.5 & 2.7  &11.5\\
  Full-IMRAM&\textbf{10.2} &\textbf{27.7}  &\textbf{3.4} &\textbf{21.7}  \\
  \hline
\end{tabular}}
\end{table}

\subsection{Results on the Newly-Collected Ads Dataset}
We evaluate our proposed IMRAM on our \Ads{} dataset. We compare our models with the state-of-the-art SCAN models in~\cite{lee2018stacked}. Comparison results are shown in Table~\ref{tab:ads_res}. We can see that the overall performance on this dataset is greatly lower than those on benchmark datasets, which indicates the challenges of cross-modal retrieval in real-world business advertisement scenarios. Results also show that our models can obtain substantial improvements over compared models, which demonstrates the effectiveness of the proposed method in this dataset.


\section{Conclusion}
In this paper, we propose an Iterative Matching method with a Recurrent Attention Memory network (IMRAM) for cross-modal image-text retrieval to handle the complexity of semantics. Our IMRAM can explore the correspondence between images and texts in a progressive manner with two features: (1) an iterative matching scheme with a cross-modal attention unit to align fragments from different modalities; (2) a memory distillation unit to refine alignments knowledge from early steps to later ones. We validate our models on three benchmarks (\ie{} Flickr8K, Flickr30K and MS COCO) as well as a new dataset (\ie{} \Ads{}) for practical business advertisement scenarios. Experiment results on all datasets show that our IMRAM outperforms compared methods consistently and achieves state-of-the-art performance.

{\small
\bibliographystyle{ieee_fullname}
\bibliography{egbib}
}

\end{document}